\documentclass[review]{elsarticle}
\usepackage{lineno,hyperref}
\usepackage[T1]{fontenc}
\usepackage{times}  
\usepackage{helvet}  
\usepackage{courier}  
\usepackage{url}  
\usepackage{graphicx}  
\usepackage{amsmath}
\usepackage{algorithm}
\usepackage{algorithmic}
\usepackage{csquotes}
\usepackage{color}
\usepackage{paralist}
\usepackage{amssymb}
\usepackage{indentfirst}
\usepackage{subfigure}
\usepackage{float}
\usepackage{multirow}
\usepackage{mathrsfs}
\modulolinenumbers[5]
\journal{Journal of \LaTeX\ Templates}

\begin{document}

\begin{frontmatter}

\title{Quality-Aware Multimodal Saliency Detection via Deep Reinforcement Learning}

\tnotetext[mytitlenote]{The authors are with the School of Computer Science and Technology, Anhui University, Hefei 230601, China. The first two authors contribute equally to this paper. E-mail: wangxiaocvpr@foxmail.com.}

\author{Xiao Wang*, Tao Sun*, Rui Yang, Chenglong Li, Bin Luo, Jin Tang}
\address[mymainaddress]{School of Computer Science and Technology, Anhui University, Hefei, Anhui Province, China}

\begin{abstract}
	Incorporating various modes of information into the machine learning procedure is becoming a new trend. And data from various source can provide more information than single one no matter they are heterogeneous or homogeneous. Existing deep learning based algorithms usually directly concatenate features from each domain to represent the input data. Seldom of them take the quality of data into consideration which is a key issue in related multimodal problems. In this paper, we propose an efficient quality-aware deep neural network to model the weight of data from each domain using deep reinforcement learning (DRL).  Specifically, we take the weighting of each domain as a decision-making problem and teach an agent learn to interact with the environment. The agent can tune the weight of each domain through discrete action selection and obtain a positive reward if the saliency results are improved. The target of the agent is to achieve maximum rewards after finished its sequential action selection. We validate the proposed algorithms on multimodal saliency detection in a coarse-to-fine way.  The coarse saliency maps are generated from an encoder-decoder framework which is trained with content loss and adversarial loss. The final results can be obtained via adaptive weighting of maps from each domain. Experiments conducted on two kinds of salient object detection benchmarks validated the effectiveness of our proposed quality-aware deep neural network.
\end{abstract}


\end{frontmatter}


\section{Introduction} 

Computer vision have achieved great success based on deep learning techniques which rely on large scale training data in recent years. Some tasks focus on dealing with single modal data ( \emph{e.g.} RGB, depth or thermal image/videos), such as visual tracking, salient object detection, which are easily influenced by illumination, clutter background, \emph{etc}.  Recent works validated the effectiveness of incorporating various modal information no matter these data are heterogeneous or homogenous (named multi-modal). For example, RGB-Depth or RGB-Thermal data is introduced into the multi-modal moving object detection \cite{Li2017Weighted}, visual tracking \cite{Li2016Learning} or salient object detection \cite{Li2017A}; and RGB-Text-Speech is introduced into sentiment analysis \cite{poria2016fusing}. Although a good deep neural network already can be obtained by these data in corresponding task, however, the model may still works not well when challenging factors occurred. 
	
According to our observations, the data collected from different domains can be complementary to each other. However, in some cases, there are only limited modals can provide useful information for the training of deep neural networks. If these data are treated equally, the noisy modal will mislead the final representation. To make deep neural network robust to modality with poor quality as mentioned above and simultaneously use the rich information from the rest domains, a quality measure mechanism is required in the design of network architecture. 
	
The target of this paper is to optimize hyperparameters dynamically according to different quality of different domains. One intuitive idea is to employ a CNN to predict the quality-aware hyperparameters for each sequence. However, we do not have the ground truth values of the hyperparameters, thus we can not provide target object values for the network to regress them in a supervised manner. Inspired by recent progress in deep reinforcement learning, an agent is trained to learn a policy to take a better action by giving a reward for its action according to the current state. The learning goal is to maximize the expected returns in a time sequence, where the return at each time step is defined as the summed rewards from this time step to the end of sequence. For our quality-aware multi-modal task, we utilize a neural network to represent the agent and allow it choose the quality weights for each image by regarding the choice as its action. By defining the reward as saliency detection accuracy, the goal of reinforcement learning becomes maximizing the expected cumulative salient object detection accuracies, which is consistent with the saliency evaluation. Similar views can also be found in \cite{Dong_2018_CVPR}. 
	
\begin{figure*}[t]
\center
\includegraphics[width=4.5in]{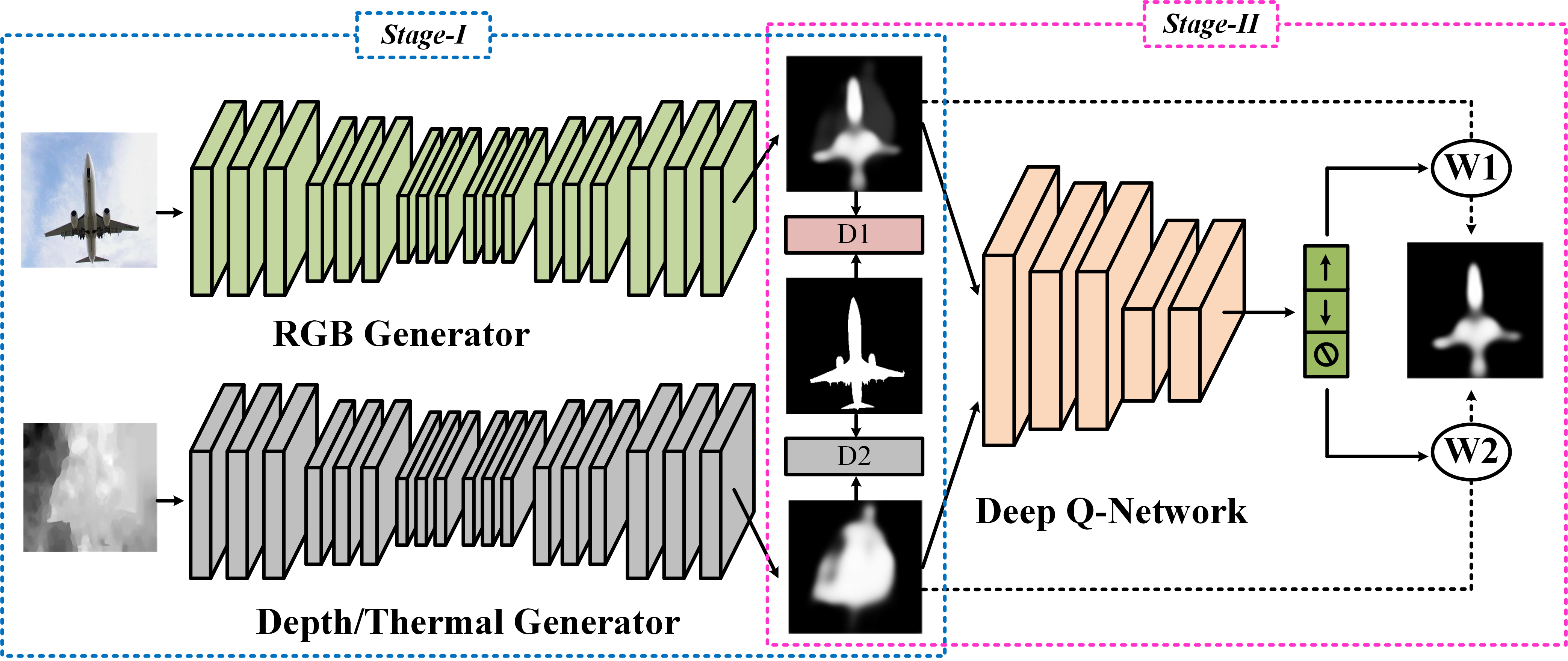}
\caption{ The pipeline of our proposed quality-aware multi-modal saliency detection network. }\label{pipeline}
\end{figure*}

In this paper, we propose a general quality estimation network which could perceive the quality of input data from different sensors, the whole pipeline can be found in Figure \ref{pipeline}. Specifically, we take the quality estimation of each domain as a decision-making problem, and train an agent to interact with the environment to explore and learn to weight each domain. The state is the input data from different domains, the actions are \emph{increase}, \emph{decrease} or \emph{terminate tune} the weight of each modality, and the reward is calculated according to the loss between estimated results and ground truth. The training of the introduced quality estimation network can be optimized by deep reinforcement learning algorithms, we adopt deep Q-network \cite{Mnih2015Human} in this paper due to it is simple and efficient to implement. More advanced reinforcement learning techniques, such as dueling network architectures for deep reinforcement learning \cite{wang2016dueling} or actor-critic algorithm \cite{mnih2016asynchronous} can also be applied in our settings. The network can automatically assign low quality scores to modality with poor quality in order to make the final results more accurate. We show the applications of the proposed quality estimation network on multi-modal salient object detection in this paper. 
	
The main contributions of this paper can be summarized as follows: 
	
$\bullet$ We introduce a novel and general quality estimation network using deep reinforcement learning which do not require any explicit annotations of the quality. 
	
$\bullet$ We apply the introduced quality estimation network on the multi-modal saliency detection task successfully, and further propose a coarse-to-fine salient object detection framework based on generative adversarial network. 
	
$\bullet$ Extensive experiments on  two public multimodal saliency detection dataset validated the effectiveness of the introduced algorithm.

\section{Related Works}
In this section, we give a brief review of multi-modal saliency detection methods, deep reinforcement learning and generative adversarial network, respectively. The comprehensive literature reviews on these saliency detection methods can be found in \cite{Peng2014RGBD} \cite{Borji2014Salient}.
	
\textbf{Deep Reinforcement Learning.} Deep learning and reinforcement learning are treated as the most important way to general artifical intelligence. Different from supervised learning and unsupervised learning, reinforcement learning target at learning to execute the "right" action in a given environment (state) and obtain the maximum rewards. The whole learning process of the agent is guided by the reward given by the environment. Deep reinforcement learning (DRL) was first proposed by Mnih \emph{et al.} \cite{Mnih2015Human} in 2013 which utilize deep neural networks, \emph{i.e.} Deep Q-learning Networks (DQN) to parametrize an action-value function to play Atari games, reaching human-level performance. The most relevant and successful application of reinforcement learning maybe the game of Go which combined policy network and value network and beat many world-class professional player \cite{Silver2016Mastering}. Asynchronous deep reinforcement learning was also introduced in \cite{Babaeizadeh2016GA3C} to tackle the training efficiency issue by Mnih \emph{et al.} On the aspect of computer vision applications, DQN also applied to many domains, such as object detection \cite{caicedo2015active, Kong2017Collaborative, Jie2017Tree},  visual tracking \cite{yun2017adnet, Wang_2018_CVPR}, Face Hallucination \cite{Cao2017Attention}.  	Caicedo \emph{et al.} introduce the DRL into the community of object detection in \cite{caicedo2015active}, and this is also the first attempt to treat the object detection task as decision-making problem. Some other DRL based object detectors futher improve the baseline algorithm by introduce tree-structured search process \cite{Jie2017Tree} or multi-agent DRL algorithm \emph{et al.}  Sangdoo \emph{et al.} \cite{yun2017adnet} propose the action-decision network to treat the visual tracking task as desion-making problem and teach the agent to learn to move the bounding box along with target object. However, there are still no prior works focus on handling the saliency detection problem with deep reinforcement learning technique. Our work is the first to introduce the DRL into the multi-modal saliency detection community to automatically learn to weight different data to better fuse the multi-modal information.

\textbf{Generative Adversarial Network.} More and more researchers focus their attention on generative adversarial networks (GANs), which is first proposed by Goodfellow in \cite{Goodfellow2014Generative}. Recently, massive works attempt to generate more realistic images \cite{Arjovsky2017Wasserstein} \cite{Gulrajani2017Improved} and also some interesting image transformation based works \cite{Isola2016Image} \cite{Dong2017Unsupervised}. Image-conditioned GAN for super-resolution which is proposed by Ledig \emph{et al.} achieved amazing performance \cite{Ledig2016Photo}. Pan \emph{et al.} first proposed to generate saliency results of given images based on GAN in \cite{Pan2017SalGAN}. GANs also achieved great success on text based image generation, such as \cite{Dash2017TAC}. Li \emph{et al.} propose to use perceptual GAN to handle the issue of small object detection in \cite{Li2017Perceptual}.  Besides, the studies about theoretical model of GAN are also one of the most hottest topic in recent years \cite{Arjovsky2017Wasserstein} \cite{Saatchi2017Bayesian} \cite{Wang2017IRGAN} \cite{Yu2016SeqGAN}. To the best of our knowledge, this work makes the first attempt to introduce GANs on the multi-modal saliency detection task.
	
\textbf{Multi-Modal Saliency Detection.} Multi-modal saliency detection discussed in this paper mainly focus on RGBD and RGBT.  Different from RGB saliency detection, multi-modal salient object detection receives less research attention \cite{Maki1996A}, \cite{Lang2012Depth}, \cite{Desingh2013Depth}, \cite{Zhang2010Stereoscopic}, \cite{Shen2012A}. An early computational model on depth-based attention by measuring disparity, flow and motion is proposed by Maki \emph{et al.} \cite{Maki1996A}. Similarly, Zhang \emph{et al.} propose a stereocopic saliency detection algorithm on the basis of depth and motion contrast for 3D videos in \cite{Zhang2010Stereoscopic}. Desingh \emph{et al.} \cite{Desingh2013Depth} estimate saliency regions by fusing the saliency maps produced by appearance and depth cues independently. However, these methods either treat the depth map as an indicator to weight the RGB saliency map \cite{Maki1996A}, \cite{Zhang2010Stereoscopic} or consider depth map as an independent image channel for saliency detection \cite{Desingh2013Depth}, \cite{Lang2012Depth}. On the other hand, Peng \emph{et al.} \cite{Peng2014RGBD} propose a multi-stage RGBD model to combine both depth and appearance cues to detect saliency. Ren \emph{et al.} \cite{Ren2015Exploiting} integrate the normalized depth prior and the surface orientation prior with RGB saliency cues directly for the RGBD saliency detection. These methods combine the depth-induced saliency map with RGB saliency map either directly  \cite{Ju2014Depth}, \cite{Ren2015Exploiting} or in a hierarchy way to calculate the final RGBD saliency map \cite{Peng2014RGBD}. However, these saliency map level integration is not optimal as it is restricted by the determined saliency values.

\section{Our Method}
	
	In this section, we will first give an overview of the designed quality-aware multi-modal saliency detection networks and the whole pipeline can be found in Figure \ref{pipeline}. Then, we will introduce the coarse single modal saliency estimation network. After that, we will give a detailed explanation about why and how to adaptively weighting the multimodal data via deep reinforcement learning. Finally, we will talk about how to train and test the adaptive weighting module.

\subsection{Overview}
	
	To validate the effectiveness of our proposed general quality estimation network, we implement our experiments based on the multi-modal saliency detection. This task target at handling the problem of finding the salient regions from multi-modal data. And the key of this task lies on how to adaptively fuse the multi-modal data to predict the final saliency results. The proposed multi-modal saliency detector dynamically pursues the target by adaptively weight each saliency results using deep reinforcement learning as shown in Figure \ref{pipeline}. 
	
	For the coarse single modal saliency estimation network, we introduce the conditional generative adversarial network (CGAN) to predict coarse saliency maps. The CGAN consists of two sub-networks, \emph{i.e.} the generator G and discriminator D. The generator follows the encoder-decoder framework, specifically, the encoder is a truncated VGG network ( with the fully connected layers removed ) which is used to extract the feature of input images; the decoder is a reversed truncated VGG network which is utilized to upsample the encoded information and output its saliency detection results. The discriminator is a standard convolutional neural network (CNN), which is introduced to detect whether the given image is real ( from ground truth saliency maps ) or fake ( from generated saliency results). With the competition between these two models, both of them can alternatively and iteratively boost their performance. Moreover, we also adopt the content loss to stable the training of GAN and speed up the training process as \cite{bousmalis2017unsupervised} \cite{Pan2017SalGAN} does. Hence, for each modal, we have one coarse saliency results produced by corresponding saliency generation network.
	
	We deal with the adaptive fusion mechanism using deep reinforcement learning which can fuse the multi-modal data through the interaction between the agent and environment. We denote the output of GANs as state, the increase, decrease or terminate the tuning of weight values are actions, and we give the agent a positive/negative reward according to the loss between predicted saliency maps and the ground truth. In the testing phase, the deep Q-network can be directly used to predict the weight of each modal until the \emph{trigger} action selected or other conditions are met. This is the first time to take the quality-aware multi-modal adaptive fusion as decision making problem and the proposed weighting mechanism can also be applied in other quality-aware tasks.

	\subsection{Review: Generative Adversarial Network}
	GANs attempt to learn a mapping from random noise vector \emph{z} to generated image \emph{y:} $z \rightarrow y$ in an unsupervised way \cite{Goodfellow2014Generative}. They utilize a discriminative network D to judge one sample comes from the dataset or produced by a generative model G. These two networks \emph{i.e.} G and D are simultaneously trained so that G learns to generate images that are hard to classify by D, while D attempt to discriminate the images generated by G. Finally, it is not easy for D to detect when G is well trained.

	The whole training procedure of GANs can be regarded as a \emph{min-max} process:
\begin{equation}
\label{GANLearnObject}
\begin{aligned}
	\mathcal{L}_{GAN}(G, D) =
  & \mathbb{E}_{y \sim P_{data} (y)} [log D(y)] + \\
  & \mathbb{E}_{x \sim P_{data} (x), z \sim P_{z}(z)} [log(1 - D(G(x, z))]
\end{aligned}
\end{equation}

	Conditional GANs generate images \emph{y} based on random noise vector \emph{z} and observed image \emph{x:} $\{ x, z \} \rightarrow y$. The whole training procedure of CGANs can be formulated as:
\begin{equation}
\label{CGANLearnObject}
\begin{aligned}
	\mathcal{L}_{cGAN}(G, D) =
  & \mathbb{E}_{x, y \sim P_{data} (x, y)} [log D(x, y)] + \\
  & \mathbb{E}_{x \sim P_{data} (x)} [log(1 - D(x, G(x, z)))]
\end{aligned}
\end{equation}
	
	D. Pathak \emph{et al.} found that the combination of CGANs and traditional loss such as \emph{$L_1$  loss} will generate more realistic images in \cite{Pathak2016Context}. The job of discriminator keep unchanged, however, the generator not only try to fool the discriminator, but also need to fit the given ground truth in an $L_1$ sense:
\begin{equation}
\label{L1Loss}
\mathcal{L}_{L1} (G) = \mathbb{E}_{x, y \sim P_{data} (x, y), z \sim P_{z}(z)} [|| y - G(x, z) ||_1] 	
\end{equation}

\subsection{Network Architecture}	
	
	As shown in Figure \ref{pipeline}, our multi-modal saliency detection can be divided into two main stages. In the stage-I, we take the multi-modal data as our input and directly output corresponding coarse saliency maps. To achieve this target, we introduce the encoder-decoder architecture which contain two truncated VGG networks. This encoder-decoder architecture has been widely used in many tasks, especially in semantic segmentation \cite{badrinarayanan2017segnet}, saliency detection \cite{Zhao2015Saliency}, \emph{etc}. Specifically, we remove the fully connected layers from standard VGG network as encoder and reverse the network as the decoder network. Hence, we can obtain coarse saliency maps from these sub-network. The weight parameter of encoder is initialized with weights of the VGG-16 model which is first pre-trained on the ImageNet dataset for general object classification \cite{Deng2009ImageNet}. The weights for the decoder are randomly initialized. In the training phase, we fix the parameter of earlier layers and only fine-tuning the last two groups of convolutional layers in VGG-16 for saving computational resources. We set the discriminator as the same with \cite{Pan2017SalGAN}, which composed of six 3$\times$3  kernel convolutions interspersed with three pooling layers, and followed by three fully connected layers.
	
\begin{table}[htp!]
\center
\caption{Detailed Configurations of Discriminator of the Generative Adversarial Network.}\label{configurationDiscriminator}
\begin{tabular}{l|cccc|c}
\hline
\hline
layer		&depth		&kernel			&stride		&pad			&activation \\
\hline
conv1-1	&3				&1$\times$1		&1				&1				&ReLU  \\
conv1-2	&32			&3$\times$3		&1				&1				&ReLU \\
pool1		& -			&2$\times$2  	&2				&0				&- \\
\hline
conv2-1	&64			&3$\times$3		&1				&1				&ReLU\\
conv2-2	&64			&3$\times$ 3		&1				&1				&ReLU\\
pool2		& -				&2 $\times$2  	&2				&0				&-\\
\hline
conv3-1	&64			&3$\times$ 3		&1				&1				&ReLU\\
conv3-2	&64			&3 $\times$3		&1				&1				&ReLU\\\
pool3		& -				&2$\times$ 2  	&2				&0				&-\\
\hline
fc4			&100			&-	 		&-			&-				&tanh\\
fc5		&2				&- 		&-			&-				&tanh\\
fc6			&1 			&-  		&-			&-  			&sigmoid	\\
\hline
\end{tabular}
\end{table} 	
	
	How to adaptively fuse these coarse results is another key problem in multi-modal tasks. The target of this paper is attempt to optimize hyperparameters dynamically according to different quality of different domains (in this paper, i.e. the RGB and thermal images, RGB and depth images). One intuitive idea is to employ a CNN to estimate the quality-aware hyperparameters for each sequence. However, we do not have the ground truth values of these parameters, therefore, we can not provide target object values for the network to train in the popular supervised way. Motivated by recent development in deep reinforcement learning, we treat these results as state and train an agent to interact with the environment to capture the quality of input data for better information fusion. This will work due to the observation that the learning target is to maximize the expected returns in a time sequence, where the return at time step t is defined as the accumulation of rewards from t to the end of the sequence. For our quality-aware multi-modal task, we utilize a neural network to represent the agent and allow it choose the quality weights for each image by regarding the choice as its action. By defining the reward as saliency detection accuracy, the goal of reinforcement learning becomes maximizing the expected cumulative salient object detection accuracies, which is consistent with the saliency evaluation. Similar views can also be found from \cite{Dong_2018_CVPR}. 
	
	The goal of agent is to give a suitable weight variable for each modal data that can be learned from the environment. During the training phase, the agent receives positive and negative rewards for each decision made when interacting with the environment. When testing, the agent does not receive any rewards and does not update the model either, it just follows the learned policy. Formally, the Markov Decision Process (MDP) has a set of actions A, a set of states S, and a reward function R. And we define these basic elements as follows:
	
	$\textbf{State.}$ The state of our agent is actually a tuple $S = \{s1, s2, s3\}$ which contain three main components, \emph{i.e.} the coarse saliency results $s1, s2$ from each subnetwork, the fused results $s3$ in previous steps. We resize and concatenate these three results into a tensor whose dimension is $56*56*3$ as our state and input to subsequent two fc layers to output the actions.

	$\textbf{Action.}$ We design three actions to adjust the weights which can be divided into two streams, \emph{i.e. adjust the weights} and \emph{terminate the adjust}. The agent can select a series of actions (\emph{i.e. increase} or \emph{decrease}) to tune the weight and finally select the \emph{terminate} action to achieve the goal of automatic weighting on the basis of the input state. The initial weight value for each modal is $1/M$. 
	
	$\textbf{Reward.}$ The target of agent is to obtain the maximum rewards, thus, the design of reward will be key to the success of learned policy. And it can be estimated during the training phase only because it requires ground truth saliency maps to  be calculated. In this paper, we utilize the fused final saliency results as the criterion of rewards. We assume the mean squared loss between the predicted salient object and ground truth saliency maps is $\mathcal{L}_{MSE}$. The reward for the \emph{increase/decrease actions} can be setted as:
\begin{equation}
\label{rewardFunction}
 R_a(s, s')=
\left\{
\begin{aligned}
\ &	+1, ~~~ if~~~ \mathcal{L}_{MSE}^{current} - \mathcal{L}_{MSE}^{prev} < 0 \\
 &	-1, ~~~ else \\
\end{aligned}
\right.
\end{equation}
where $s$ and $s'$ are current and next state, respectively. This equation means that we will give a positive reward if the loss decreased after a series of weighting tuning. Otherwise, we will punish the agent by giving a negative reward.

	When to stop this adjust process is another key point to the success of adaptive weighting mechanism. Because maybe we can not obtain the optimal weights, if the adjustment stopped too early. On the other hand, the time consuming will be large, if we can not timely stop the operations. Hence, we designed another specified reward function for the \emph{terminate action}:
\begin{equation}
R_t(s, s')=\left\{
\begin{aligned}
\ &	+ \eta, ~~~ if~~~ \mathcal{L}_{MSE} \leq \phi \\
  &	- \eta, ~~~ else
\end{aligned}
\right.
\end{equation}
where $\phi$ is a pre-defined threshold parameter (we set $\phi$ as 0.04 in our experiments). This function denotes that if the agent choose the \emph{terminate action}, we will compute the final weighted saliency results and compare it with ground truth saliency maps to obtain the MSE value of current state. If the value of MSE is less than the given threshold $\phi$, we think it's time to stop the weight adjustment and give a positive reward $+\eta$ to the agent, otherwise, we give a negative $\eta$ to punish the agent.

\subsection{The Training}	
	
	We train the quality-aware multi-modal saliency detection network into two stages. We first train the coarse saliency estimation network with mean squared loss and adversarial loss in stage-I. Then, we train the adaptive fusion module (\emph{i.e.} the deep reinforcement learning) in the stage-II. The loss funcations used in these two stages are introduced as follows respectively.

	
	\textbf{Loss Function in Stage-I.} To achieve better saliency estimation, the proposed encoder-decoder architecture is trained by combining a content loss and adversarial loss which has been widely used in many prior works \cite{Ledig2016Photo} \cite{Pan2017SalGAN}. Content loss is computed in a per-pixel basis, where each value of the predicted saliency map is compared with its corresponding peer from the ground truth map. Assume we have an image $I$ and its resolution is $N=W \times H$, and the ground truth saliency maps can be denoted as $S$, the predicted saliency maps is $\hat{S}$. The content loss which measures the mean squared error (MSE) or Euclidean loss between the predicted and ground truth saliency maps can be defined as: 
\begin{equation}
\label{mseObject}
\begin{aligned}
	\mathcal{L}_{MSE} = \frac{1}{N} \sum_{j=1}^{N} (S_j - \hat{S_j})^2
\end{aligned}
\end{equation}
	
	The adversarial loss function is adopted from conditional generative adversarial networks (CGANs). This network consists of one generator and one discriminator, and these two models play a game-theoretical \emph{min-max} game. Specifically, the generative model tries to fit saliency distribution provided by reference images and produces ``fake'' samples to fool the discriminative model, while the discriminative model tries to recognize whether the sampled image is from ground truth or estimated by the generative model. With the competition between these two models, both of them can alternately and iteratively boost their performance. The mathematical function can be formulated as: 	
\begin{equation}
\label{mseObject}
\begin{aligned}
	\mathcal{L}_{GAN} = - log D(I, \hat{S})
\end{aligned}
\end{equation}
where $D(I, \hat{S})$ is the probability of fooling the discriminator, so that the loss associated to the generator will grow more when chances of fooling the discriminator are lower.

	As illustrated in above sections, we combine the MSE loss with adversarial loss to obtain more stable and fast convergence generator. The final loss function for the generator during adversarial training can be formulated as:
\begin{equation}
\label{finalObject}
\begin{aligned}
	\mathcal{L}_{total} = \mathcal{L}_{MSE} + \lambda * \mathcal{L}_{GAN}
\end{aligned}
\end{equation}
where $\lambda$ is the tradeoff parameters to balance these two loss functions. We experimentally set this parameter as 0.33 to obtain better saliency detection results.
	
	During the adversarial training, we alternate the training of the generator and discriminator after each iteration (batch). $L_2$ weight regularization (\emph{i.e.} weight decay) when training both the generator and discriminator ($\lambda = 1 \times 10^{-4}$). AdaGrad was utilized for model optimization, with an initial learning rate of $3 \times 10^{-4}$.

	\textbf{Loss Function in Stage-II.} 
	The parameters of the Q-network are initialized randomly. The agent is setted to interact with the environment in multiple episodes, each representing a different training image. We also take a $\epsilon$-greedy \cite{bellemare2013arcade} to train the Q-network, which gradually shifts from exploration to exploitation according to the value of $\epsilon$. When exploration, the agent selects actions randomly to observe different transitions and collects a varied set of experience. During exploitation, the agent will choose actions according to the learned policy and learns from its own successes and mistakes.
	
	The utilization of \emph{target network} and \emph{experience replay} \cite{lin1993reinforcement} in DQN algorithm is the key ingredient of their success. The target network with parameters $\theta^-$ is copied every $\tau$ steps from online network and kept fixed on all other steps, thus, we could have $\theta^-_i = \theta_i$. The target in DQN can be described as the following formulation:
\begin{equation}
\label{targetY}
Y^{DQN}_i \equiv  r + \gamma max_{a'} Q(s', a'; \theta_i^-)
\end{equation}
A replay memory is used to store the experiences of past episodes, which allows one transition to be used in multiple model updates and breaks the short-time strong correlations between training samples.  Each time Q-learning update is applied, a mini batch randomly sampled from the replay memory is used as the training samples. The update for the network weights at the $i^{th}$ iteration $\theta_i$ given transition samples ($s, a, r, s'$) is as follows:
\begin{equation}
\small
\theta_{i+1} = \theta_i + \alpha (r + \gamma max_{a'} Q(s', a'; \theta_i) - Q(s, a; \theta_i))\nabla_{\theta_i}Q(s, a; \theta_i).
\end{equation}
where $a'$ represents the actions that can be taken at state $s'$, $\alpha$ is the learning rate and $\gamma$ is the discount factor.

The pseudo-code for training the quality estimation network can be found in Algorithm \ref{algorithm}.

\renewcommand{\algorithmicrequire}{\textbf{Input:}}
\renewcommand{\algorithmicensure}{\textbf{Output:}}
\begin{algorithm}[htbp]
	\caption{ The Training of Quality Estimation Network. }
	\label{algorithm}
	\begin{algorithmic}[1]
		\REQUIRE Coarse RGB saliency results, Coarse depth/thermal saliency results \\
		\STATE Initialize replay memory $\mathcal{D}$ to capacity $\mathcal{N}$
		\STATE Initialize action-value function ${Q}$ with random weights $\theta$
		\STATE Initialize target action-value function $\hat{Q}$ with weights $ \theta^- = \theta $
	
		\FOR{   episode = $1, M$  }
				\FOR{ each image }
				\STATE  Initialise sequence $s_1 $ and pre-processed sequence $\phi_1 = \phi(s_1)$
	    	\FOR{ step $ t = 1, T $ }
 		
 				 	\IF{random number $\delta$ $< \epsilon$}
 						\STATE  select a random action $a_t$
 			 		\ELSE
 						\STATE   select $a_t = \arg \max_a Q(\phi(s_t); a; \theta) $
 					\ENDIF
					
					\STATE  Execute action $a_t$ to change the weight $w_{t}$ of each modal and observe reward $r_t$ and new weight $w_{t+1}$
					\STATE  Set $s_{t+1} = s_t$  and pre-process $\phi_{t+1} = \phi(s_{t+1})$
					\STATE  Store transition $(\phi_t, a_t, r_t, \phi_{t+1})$ in $\mathcal{D}$
					\STATE  Sample random mini-batch of transitions $(\phi_j, a_j, r_j, \phi_{j+1})$ from $\mathcal{D}$

 					\IF{episode terminates at step $j+1$}
 						\STATE $y_j = r_j $
 			 		\ELSE
 						\STATE $y_j = r_j + \gamma \max_{a'} \hat{Q} (\phi_{j+1}, a'; \theta^-)$
 					\ENDIF
 					
					\STATE Perform a gradient descent step on $(y_j - Q(\phi_j, a_j; \theta))^2$  with respect to the network parameters $\theta$
					\STATE reset $\hat{Q} = Q$ for every C steps
					
			\ENDFOR
 		\ENDFOR
 		 		\ENDFOR
	\end{algorithmic}
\end{algorithm}

\section{Experiments}
	In this section, we validate the proposed approach on two public multi-modal saliency detection benchmarks, including RGB-Depth (RGBD) and RGB-Thermal (RGBT) salient object detection benchmarks. We will first give an introduction about evaluation criterion and dataset description, then we will analyse the experimental results on RGBD and RGBT datasets. We also give an ablation study on the components and efficiency analysis.

\begin{table*}[t]
\tiny
\centering
\caption{Precision, Recall, F-measure of our method against different kinds of baseline methods on the public RGBD benchmark. The code type is also presented. The bold fonts of results indicate the best performance. }\label{RGBDcomparison}
\begin{tabular}{|l||c|c|c||c|c|c||c|c|c||c|c|}
\hline
\multirow{2}{*}{\textbf{Algorithm}} & \multicolumn{3}{c||}{\textbf{Color}} & \multicolumn{3}{c||}{\textbf{Depth}} & \multicolumn{3}{c||}{\textbf{Color-Depth}} & \multirow{2}{*}{\textbf{Code Type}} \\
\cline{2-10}
							& P & R & F 	& P & R &F 			& P & R & F &  \\
\hline
RR (CVPR2015)		  	 &0.7159 	 &0.5728 	 &0.6468 	 		&0.7568 	&0.7069 	&0.7175 	 		&0.7802	&0.6924	 &0.7284 			&Matlab 	 	\\
\hline
MST (CVPR2016) 		 &0.6856 	 &0.5980 	 &0.6312 	 		&0.5601 	&0.5178 	&0.5242 	  		&0.6415	&0.6276	 &0.6103   		&Wrapping code  	\\
\hline
BSCA (CVPR2015)	 	 &0.7003 	 &0.6052 	 &0.6498 	 		&0.7541 	&0.6563 	&0.6986 			&0.7542	&0.6608	 &0.7033  		&Matlab 	 	\\
\hline
DeepSaliency (TIP2016) &0.8311 	 &0.7877 	 &0.8021 	 		&0.6445 	&0.5441 	&0.5890 	 		&0.7936	&0.7656	 &0.7619  		&Caffe 	\\
\hline
DRFT (CVPR2013)	 	 &0.7254 	 &0.6164 	 &0.6668 	 		&0.6020 	&0.5327 	&0.5351 	 		&0.7678	&0.6613	 &0.6963   		&Matlab	  	\\
\hline
DSS (CVPR2017)  	  	 &0.8512 	 &0.7934 	 &0.8208 	 		&0.7021 	&0.5896 	&0.6453 			&0.8258	&0.8036	 &0.8017   		&Caffe  		\\
\hline
HSaliency (CVPR2013) &0.7048 	 &0.4820 	 &0.5891 	 		&0.3900 	&0.4547 	&0.3755 			&0.6479	&0.4991	 &0.5487   		&Wrapping code 	 	\\
\hline
MDF (CVPR2015)  	 	 &0.7845 	 &0.6584 	 &0.7153 	 		&0.5873 	&0.4882 	&0.5126 	 		&0.7696	&0.7026	 &0.7106  		&Caffe 	 	\\
\hline
RBD (CVPR2014)  	 	 &0.6537 	 &0.5771 	 &0.6053 	 		&0.5823 	&0.4967 	&0.5331 	 		&0.6607	&0.6279	 &0.6273  		&Matlab 	 	\\
\hline
MCDL (CVPR2015)	 	 &0.7516 	 &0.6009 	 &0.6739  		&0.4973 	&0.4183 	&0.4298 	   		&0.7642 &0.5088	 &0.6069   		&Matlab  		\\
\hline
MLNet (ICPR2016) 	 	 &0.6673 	 &0.3856 	 &0.5321 			&0.3349 	&0.1977 	&0.2651 	 		&0.5288	&0.3085	 &0.4207   		&Keras  	\\
\hline

Ours (Equal Weights)	&0.8407 	 &0.8575	 &0.8339 	  		&0.8312	&0.8521 	&0.8252 	 		&0.8480	&0.8625	 &0.8362    		&Theano+Lasagne 	 	\\

\hline 
Ours (Adaptive Weights)	&0.8407 	 &0.8575 	 &0.8339 	 	&0.8312 	&0.8521 	&0.8252   	&0.8541	&0.8596	 &0.8440   	&Theano+Lasagne 	 	\\
\hline

\end{tabular}
\end{table*}

\begin{table*}[t]
\tiny
\centering
\caption{Precision, Recall, F-measure of our method against different kinds of baseline methods on the public RGBT benchmark. The code type and runtime are also presented. The bold fonts of results indicate the best performance. }\label{RGBTcomparison}
\begin{tabular}{|l||c|c|c||c|c|c||c|c|c||c|c|}
\hline
\multirow{2}{*}{\textbf{Algorithm}} & \multicolumn{3}{c||}{\textbf{Color}} & \multicolumn{3}{c||}{\textbf{Thermal}} & \multicolumn{3}{c||}{\textbf{Color-Thermal}} & \multirow{2}{*}{\textbf{Code Type}} & \multirow{2}{*}{\textbf{FPS}} \\
\cline{2-10}
& P & R & F  	& P & R & F 		& P & R & F   & & \\
\hline
BR (ECCV2010) 	 	&0.724 &0.260 &0.411   		&0.648 &0.413 &0.488     		&0.804 &0.366 &0.520    		&Matlab \& C++ &8.23  \\
\hline
SR (JV2009) 	  			&0.425 &0.523 &0.377   		&0.361 &0.587 &0.362   		&0.484 &0.584 &0.432 		    	&Matlab &1.60 \\
\hline
SRM (CVPR2007) 		&0.411 &0.529 &0.384   		&0.392 &0.520 &0.380   		&0.428 &0.575 &0.411 		    	&Matlab 	&0.76	\\
\hline
CA (CVPR2015) 		&0.592 &0.667 &0.568   		&0.623 &0.607 &0.573    		&0.648 &0.697 &0.618  		&Matlab &1.14	\\
\hline
MCI (TPAMI2012) 		&0.526 &0.604 &0.485  		&0.445 &0.585 &0.435  		&0.547 &0.652 &0.515   		&Matlab\&C++ 	&\textbf{21.89} \\
\hline
NFI (JV2013) 	  	 	&0.557 &0.639 &0.532   		&0.581 &0.599 &0.541  		&0.564 &0.665 &0.544  		&Matlab &12.43	\\
\hline
SS-KDE (SCIA2011)    &0.581 &0.554 &0.532  		&0.510 &0.635 &0.497  		&0.528 &0.656 &0.515  		& Matlab\&C++ &0.94	\\
\hline
GMR (CVPR2013) 	  	&0.644 &0.603 &0.587  		&0.700 &0.574 &0.603   		&0.694 &0.624 &0.615  		&Matlab &1.11 \\
\hline
GR (SPL2013) 	  		&0.621 &0.582 &0.534   		&0.639 &0.544 &0.545   		&0.705 &0.593 &0.600   		&Matlab\&C++ &2.43 \\
\hline
STM (CVPR2013) 	  	&0.658 &0.569 &0.581  		&0.647 &0.603 &0.579  		&-		&-		&-	    				&C++ &1.54 \\
\hline
MST (CVPR2016)	  	&0.627 &0.739 &0.610  		&0.665 &0.655 &0.598  		&-		&-		&-		 			& C++ 	&0.53 \\
\hline
RRWR (CVPR2015) 	 &0.642 &0.610 &0.589   		&0.689 &0.580 &0.596  		&-		&-		&-		 			&C++ &2.99 \\
\hline
Ours (Equal Weights)      &0.8474	&0.8453 &0.8351	&0.8321	&0.8501 	&0.8251  &0.8497	&0.8595	&0.8386	 		&Python	&-	\\
\hline
Ours (Adaptive Weights) &0.8474	&{0.8453}	&{0.8351}	&{0.8321}	&{0.8501}	&{0.8251}	    &{0.8520}	 &{0.8591}	&0.8413	&Python	&5.88	\\
\hline

\end{tabular}
\end{table*}

\subsection{Evaluation Criteria and Dataset Description}
	For fair comparisons, we fix all parameters and other settings of our approach in the experiments, and use the default parameters released in their public codes for other baseline methods.  In our experiments, we set $\eta$ equal to 2 in our reward function; $\alpha$, $\gamma$ and $\epsilon$ is setting as 0.0001, 0.9, 1.0, respectively.
	
	For quantitative evaluation, we regard it as a classification problem and evaluate the results using two groups of evaluation criterion, \emph{i.e.} Precision, Recall, F-measure (P, R, F for short) and MSE. The mathematical formulations of P, R, F can be described as follows:
\begin{equation}
\label{PRF}
\begin
{aligned}
	Precision = \frac{TP}{TP+FP}; ~~~~~~
	Recall = \frac{TP}{TP+FN};  \\
	F-measure = \frac{(1+\beta)*precision * recall}{\beta*precision + recall};
\end{aligned}
\end{equation}
where TP, FP, TN and FN mean the numbers of true positives, false positives, true negatives and false negatives, respectively. We set the super-parameter $\beta$ as 0.3 in all our experiments.

	We denote the ground truth saliency map as $\hat{S}$ and the predicted results as $S$. And the mean squared error (MSE) can be written as:
\begin{equation}
\label{MSE}
\begin{aligned}
	MSE = \frac{1}{N} \sum_{t=1}^{N} (S_t - \hat{S}_t)^2 ;
\end{aligned}
\end{equation}
	
	We evaluate salient object detectors on two public saliency detection benchmarks including RGBD (named NJU2000 dataset) \cite{Ju2014Depth} and RGBT benchmarks \cite{Li2017A}. The RGBD dataset consists 2,000 stereo images, as well as corresponding depth maps and manually labeled groundtruth. These images are collected from Internet, 3D movies and photographs by a Fuji W3 stereo camera. They perfrom mask labeling in a 3D display environment by using \emph{Nvidia 3D Vision} due to the labeling results on 2D images maybe a  little different from that in real 3D environments. The project page of this benchmark can be found from this website \footnote{\url{http://mcg.nju.edu.cn/publication/2014/icip14-jur/index.html}}.
	
	To evaluate the generalization of our proposed quality-aware multi-modal deep saliency detection network, we also report the saliency detection performance on RGBT benchmark. The newest RGBT benchmark proposed by Li \emph{et al.} includes 821 aligned RGB-T images with the annotated ground truths, and it also present the fine-grained annotations with 11 challenges to allow researchers to analyse the challenge-sensitive performance of different algorithms. Moreover, they implement 3 kinds of baseline methods with different inputs (RGB, thermal and RGB-T) for evaluations. The detailed configuration of this benchmark can be found from \footnote{\url{http://chenglongli.cn/people/lcl/journals.html.}}.

\subsection{Compare with State-of-the-art Methods}
	We compare our proposed quality-aware multi-modal salient object detection network with 11 state-of-the-art saliency detectors on the RGBD saliency detection benchmark including 6 traditional methods and 5 deep learning based approaches, including: RR \cite{Li2015Robust}, MST \cite{Tu2016Real}, BSCA \cite{Qin2015Saliency}, DeepSaliency \cite{Li2015DeepSaliency}, DRFT \cite{Wang2013Salient}, DSS \cite{hou2016deeplycvpr}, HSaliency \cite{Yan2013Hierarchical}, MDF \cite{Li2015Visual}, RBD \cite{Zhu2014Saliency}, MCDL \cite{Zhao2015Saliency}, MLNet \cite{mlnet2016}.

	The baseline methods we compared on RGBT saliency detection dataset are directly adopted from this benchmark.  The saliency detection performance of our proposed method and other start-of-the-art detectors on the two benchmarks will be discussed in later subsections, respectively.

\subsubsection{Results on RGB-Depth Dataset}
	We first report the Precision, Recall and F-measure of each method on the entire dataset as shown in Table \ref{RGBDcomparison}. From the evaluation results, we can find that the proposed method substantially outperforms all baseline approaches. This comparison clearly demonstrates the effectiveness of our approach for adaptively fuse color and depth images. Besides, we can also discover that the proposed quality-aware adaptive weighted RGB-D saliency results are significantly better than single modal results. This fully demonstrate the depth images are effective to boost image saliency detection and complementary to RGB data.
	
\begin{figure}[t]
\center
\includegraphics[width=3.5in]{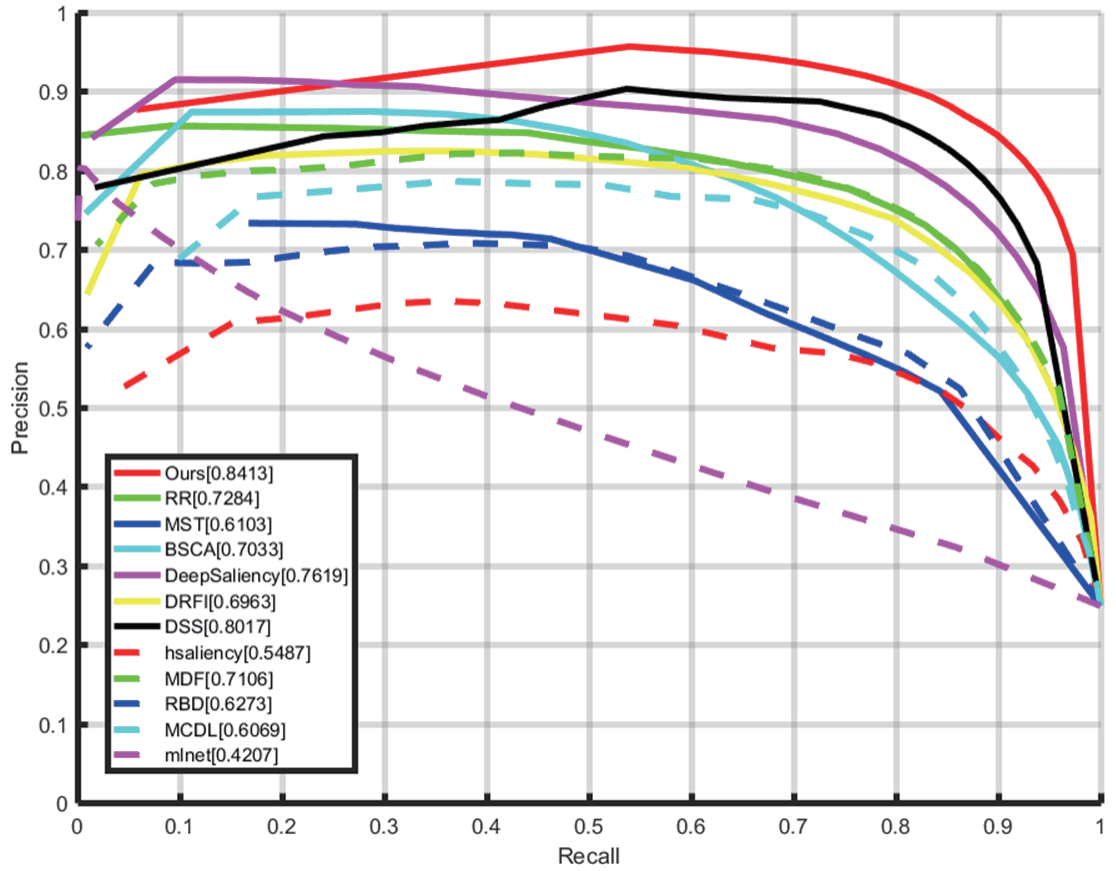}
\caption{PR curve of RGBD saliency benchmark.}\label{PR_curve_RGBD}
\end{figure}

\begin{figure*}[t]
\center
\tiny
\includegraphics[width=5in]{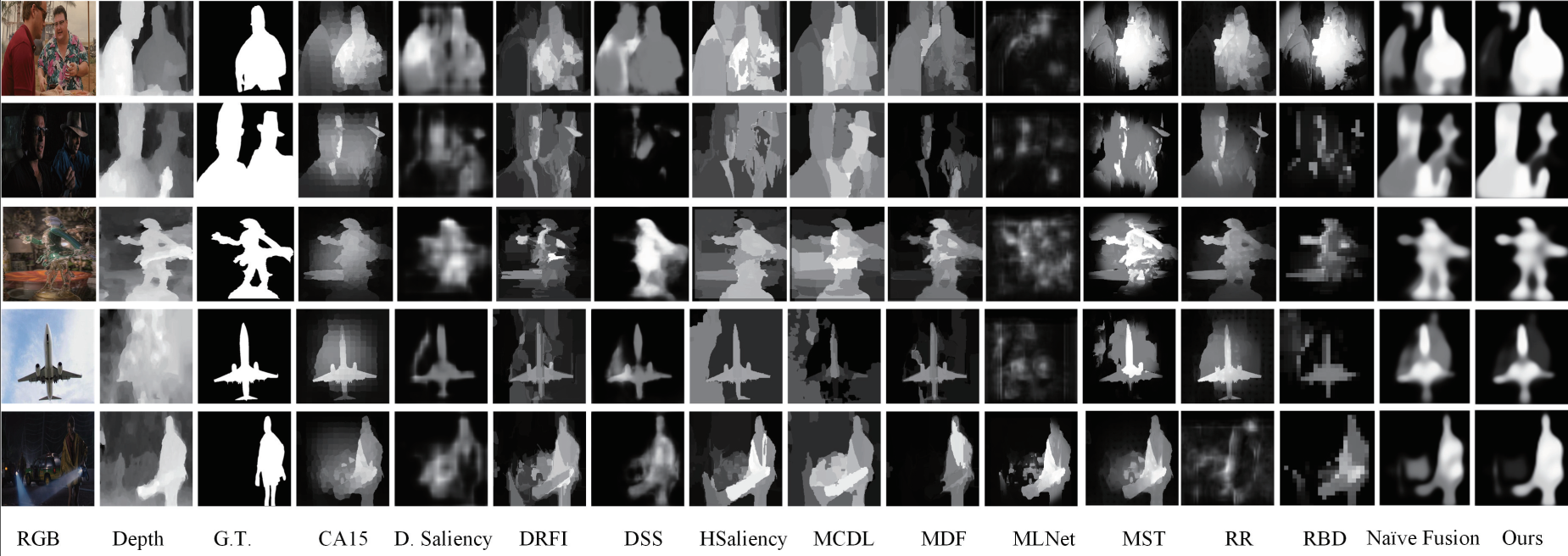}
\caption{Sample images of saliency detection results on RGBD saliency benchmark.}\label{RGBDsaliency}
\end{figure*}

	To give a more intuitive understanding of all the saliency detection results, we give a PR-curve as shown in Figure \ref{PR_curve_RGBD}. It is easy to find that the proposed method can achieve better salient object detection results compared with other state-of-the-art approaches. The saliency detection results can be found in Figure \ref{RGBDsaliency}.
	
\subsubsection{Results on RGB-Thermal Dataset}
	To further validate the generic and the effectiveness of our quality-aware deep multi-modal saliency detection network, we also implement the experiments on another multi-modal dataset, \emph{i.e.} RGB-Thermal dataset. We also report the detection results on Precision, Recall, F-measure values on this dataset. The specific saliency detection results of our and other state-of-the-art algorithms can be found in Table \ref{RGBTcomparison}. Similar conclusions can also be drawn from this dataset, and we do not reiterate them here.

\subsection{Ablation Study}
	We discuss the details of our approach by analysing the main components and efficiency in this section.
		
	\textbf{Components Analysis.} To justify the significance of the main components of the proposed approach, we implement two special versions for comparative analysis, including: 1) Ours-I, that removes the adversarial loss in the proposed network architecture, \emph{i.e.} only the MSE  loss used to train the network; 2) Ours-II, removes the modal weights and naively fuse the multi-modal data with equal contributions. Intuitively, Ours-I is designed to validate the effectiveness of adversarial training, and Our-II is implemented to check the validity of quality-aware deep Q-network which used to adaptively measure the quality of multi-modal data.
	
	As the MSE results presented in Table \ref{MSE_AS}, and we can summarize the following conclusions. 1) The complete algorithm achieves superior performance than Ours-I, validating the effectiveness of adversarial loss. 2) Our method substantially outperform Ours-II. This demonstrate the significance of the introduced quality-aware deep Q-network to achieve adaptive fusion of different source data. It is also worthy to note that the proposed quality-aware weighting mechanism is a general adaptive weighting framework and it can also be applied in many other related tasks, such as multi-modal visual tracking, multi-modal moving object detection or quality-aware procedure. We leave this for our future works.
	
\begin{table}[htp!]
\center
\caption{MSE score of the ablation study on RGBD dataset.}\label{MSE_AS}
\begin{tabular}{c|ccc}
\hline
\hline
Algorithms   	   &Ours-I		&Ours-II    & Ours 	\\
\hline
MSE			  &0.0788 		& 0.0803 		&0.0782  	\\	
\hline	
\end{tabular}
\end{table}

	\textbf{Efficiency Analysis.} Runtime of our approach against other methods are all presented in Table \ref{RGBTcomparison} (in the column FPS). The experiments are carried out on a desktop with an Intel I7 3.4GHz CPU, GTX1080 and 32 GB RAM, and our code is implemented based on the deep learning framework Theano \footnote{\url{http://deeplearning.net/software/theano/}} and Lasagne  \footnote{\url{http://lasagne.readthedocs.io/en/latest/ }}. It is obviously that our method achieved better trade-off between detection accuracy and efficiency.


\subsection{Discussion}
\begin{figure*}[t]
\center
\tiny
\includegraphics[width=3in]{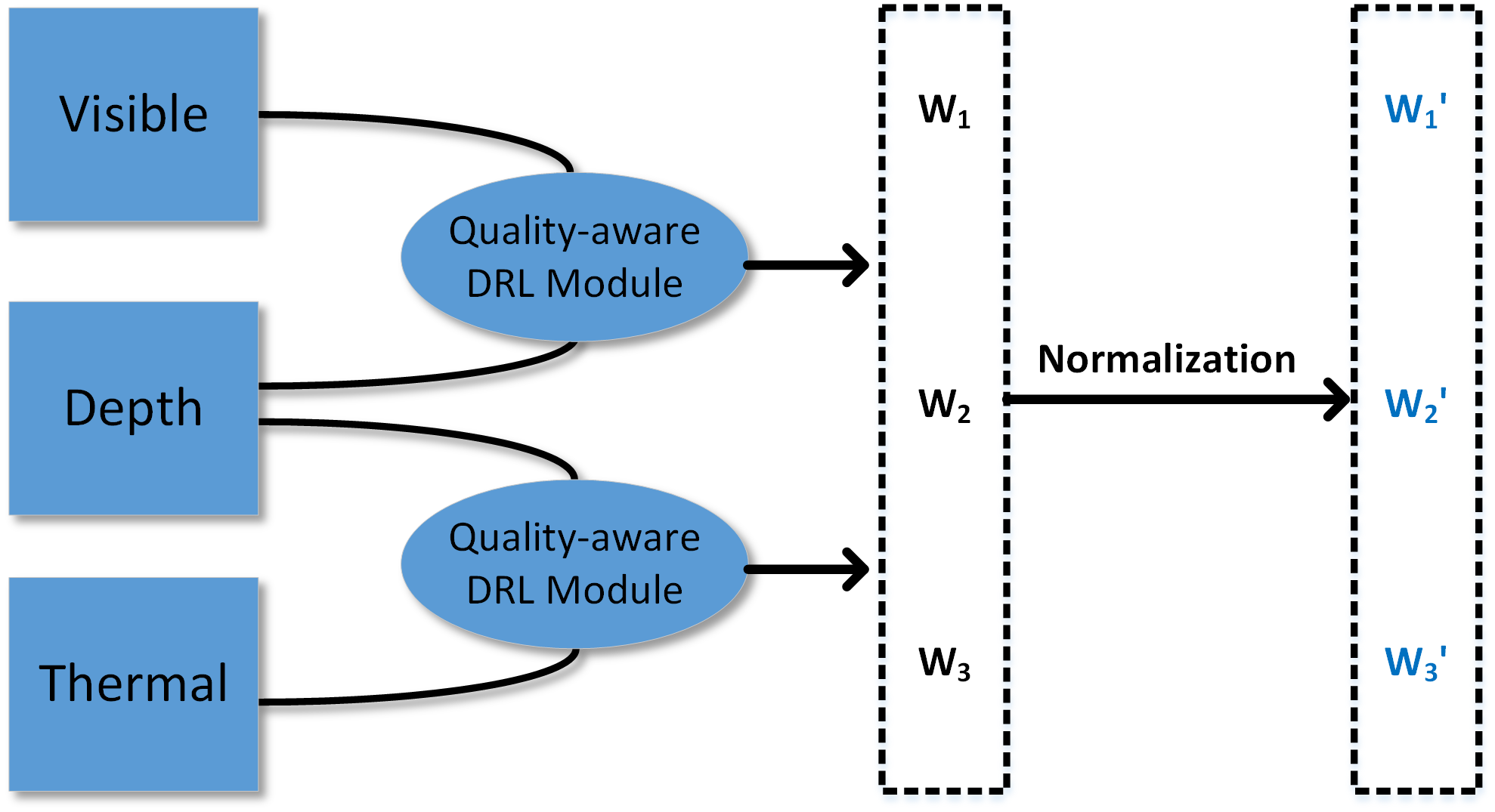}
\caption{The illustration of our approach to deal with more modalities.}\label{RGBDTsaliency}
\end{figure*}

In this paper, we validated the effectiveness of our algorithm on the task of multi-modal saliency detection. Specifically speaking, only two modality are contained in our case, \emph{i.e.} RGB-Thermal or RGB-Depth images. How to deal with more modalities with our method is also worthy to consider, for example, RGB-Thermal-Depth image pairs. As shown in Figure \ref{RGBDTsaliency}, we can adaptive weighting these modalities in a sequential manner. Another possible solution is that, we take these modalities as the input state, and output corresponding weights directly. We leave these ideas as our future works.

\section{Conclusion}
	In this paper, we propose a novel quality-aware multi-modal saliency detection neural network using deep reinforcement learning. To the best of our knowledge, this is the first attempt to introduce the deep reinforcement learning into the multi-modal saliency detection problem to handle the adaptive weighting of different modal data. Our network architecture follow the coarse-to-fine framework, that is to say, our pipeline consist of two sub-networks, \emph{i.e.} coarse single modal saliency estimation network and adaptive fusion Q-network. For each modal, we detect salient objects using encoder-decoder network and train the network with content loss and adversarial loss. We take the adaptive weighting of different data in multi-modal case as decision making problem and teach the agent to learn a weighting policy through the interaction between the agent and environment. It is also worthy to note that our adaptive weighting mechanism is a general weighting method and it can also be applied in other related tasks. Extensive experiments on RBGD and RGBT benchmarks validated the effectiveness of our proposed quality-aware deep multi-modal salient object detection network.

%
%

\section{References}
\bibliography{reference}

\end{document}